\documentclass[letterpaper, 10 pt, conference]{ieeeconf}  

\IEEEoverridecommandlockouts                              

\usepackage{amsmath} 
\usepackage{amssymb}  

\usepackage{graphicx}
\usepackage{booktabs}
\usepackage{multirow} 
\usepackage{multicol}
\usepackage[utf8]{inputenc}
\usepackage[T1]{fontenc}
\usepackage{makecell} 
\usepackage{hyperref}
\usepackage{xcolor}

\usepackage{colortbl}  

\usepackage{float} 

\title{\LARGE \bf
\underline{D}iffusion \underline{K}nows \underline{T}ransparency: Repurposing Video Diffusion for Transparent Object Depth and Normal Estimation
}

\author{
Shaocong Xu$^{1}$,
Songlin Wei$^{2}$,
Qizhe Wei$^{1}$,
Zheng Geng$^{1}$,
Hong Li$^{1,4}$,
Licheng Shen$^{3}$,
Qianpu Sun$^{3}$,
Shu Han$^{5}$\\
Bin Ma$^{3}$,
Bohan Li$^{6,7}$,
Chongjie Ye$^{8}$,
Yuhang Zheng$^{9}$,
Nan Wang$^{1}$,
Saining Zhang$^{1}$,
and Hao Zhao$^{1,3}$%
\thanks{$^{1}$Beijing Academy of Artificial Intelligence,
{\tt\small scxu@baai.ac.cn}}%
\thanks{$^{2}$University of Southern California.
}%
\thanks{$^{3}$Tsinghua University, {\tt\small zhaohao@air.tsinghua.edu.cn}.
}%
\thanks{$^{4}$Beihang University.
}%
\thanks{$^{5}$Wuhan University.
}%
\thanks{$^{6}$Shanghai Jiao Tong University.
}%
\thanks{$^{7}$European Institute of Innovation and Technology Ningbo.
}%
\thanks{$^{8}$FNii, The Chinese University of Hong Kong, Shenzhen.
}%
\thanks{$^{9}$National University of Singapore.
}%
}

\begin{document}

\maketitle
\thispagestyle{empty}
\pagestyle{empty}

\begin{abstract}

Transparent objects remain notoriously hard for perception systems: refraction, reflection and transmission break the assumptions behind stereo, ToF and purely discriminative monocular depth, causing holes and temporally unstable estimates. Our key observation is that modern video diffusion models already synthesize convincing transparent phenomena, suggesting they have internalized the optical rules. We build TransPhy3D, a synthetic video corpus of transparent/reflective scenes: 11k sequences (1.32M frames) rendered with Blender/Cycles. Scenes are assembled from a curated bank of category-rich static assets and shape-rich procedural assets paired with glass/plastic/metal materials. We render RGB + depth + normals with physically based ray tracing and OptiX denoising. Starting from a large video diffusion model, we learn a video-to-video translator for depth (and normals) via lightweight LoRA adapters. During training we concatenate RGB and (noisy) depth latents in the DiT backbone and co-train on TransPhy3D and existing frame-wise synthetic datasets, yielding temporally consistent predictions for arbitrary-length input videos. The resulting model, DKT, achieves zero-shot SOTA on real and synthetic video benchmarks involving transparency: ClearPose, DREDS (CatKnown/CatNovel), and TransPhy3D-Test. It improves accuracy and temporal consistency over strong image/video baselines (e.g., Depth-Anything-v2, DepthCrafter), and a normal variant (DKT-Normal) sets the best video normal estimation results on ClearPose. A compact 1.3B version runs at ~0.17 s/frame (832×480). Integrated into a grasping stack, DKT’s depth boosts success rates across translucent, reflective and diffuse surfaces, outperforming prior estimators. Together, these results support a broader claim: “Diffusion knows transparency.” Generative video priors can be repurposed, efficiently and label-free, into robust, temporally coherent perception for challenging real-world manipulation. Code and models are available at \url{https://daniellli.github.io/projects/DKT/}.
\end{abstract}
\section{INTRODUCTION}

Accurate depth estimation of transparent and reflective objects is fundamental to advancing 3D reconstruction \cite{grinvald2021tsdf++} and robotic manipulation \cite{cui2025gapartmanip}. Nevertheless, the intrinsic physical ambiguities of these objects impose substantial limitations on depth-sensing cameras that rely on time-of-flight measurements or stereo correspondence \cite{11128401, wei2024d}. In particular, transparent objects often produce missing regions in depth maps, which in turn lead to degraded performance in downstream tasks.

Recent data-driven approaches have sought to address this challenge by constructing datasets \cite{chen2022clearpose} that encompass diverse lighting conditions and material properties, thereby approximating the visual characteristics of transparent and specular objects, and subsequently training models for depth prediction \cite{wei2024d, shi2024asgrasp}. However, such datasets remain constrained in diversity, and the resulting methods frequently exhibit suboptimal performance in real-world scenarios. We hypothesize that these methods tend to overfit to the limited datasets on which they are trained.
\begin{figure}[t]
\includegraphics[width=1\columnwidth]{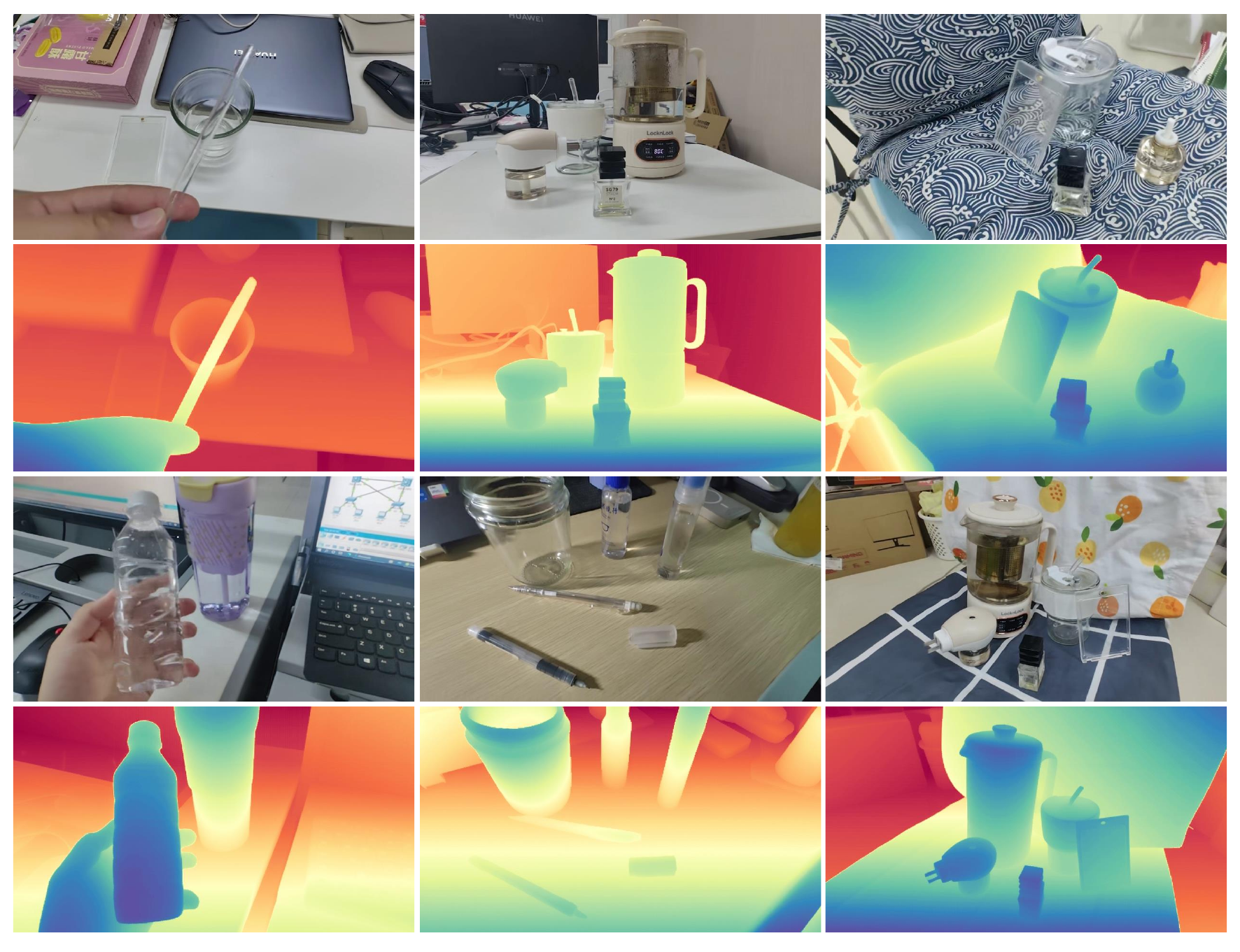}
\caption{\textbf{In-the-Wild Qualitative Results.} The first and third rows present frames extracted from the input videos, while the second and fourth rows display the predictions. Our method achieves robust depth estimation for transparent objects in arbitrary-length, in-the-wild videos. For the full video, please refer to the Appendix video.}
\label{fig:main}
\end{figure}
To address the generalization challenge, recent works \cite{wei2024d, yang2024depth} have increasingly leveraged pre-trained vision encoders, such as DINO \cite{caron2021emerging}, or harnessed text-to-image foundation models like Stable Diffusion \cite{rombach2022high} to train depth estimation networks. While these approaches have achieved notable improvements in single-frame depth accuracy, they continue to suffer from a lack of temporal consistency across frame sequences \cite{hu2025depthcrafter}. This limitation is particularly detrimental to downstream tasks that rely on stable 3D perception to support consistent action policies, such as robotic manipulation \cite{chi2023diffusionpolicy, geng2023sage}. These tasks are often carried out in dynamic and unstructured environments, where robust perception and temporally coherent decision-making are indispensable.

\begin{figure*}[th]
\includegraphics[width=2.0\columnwidth]{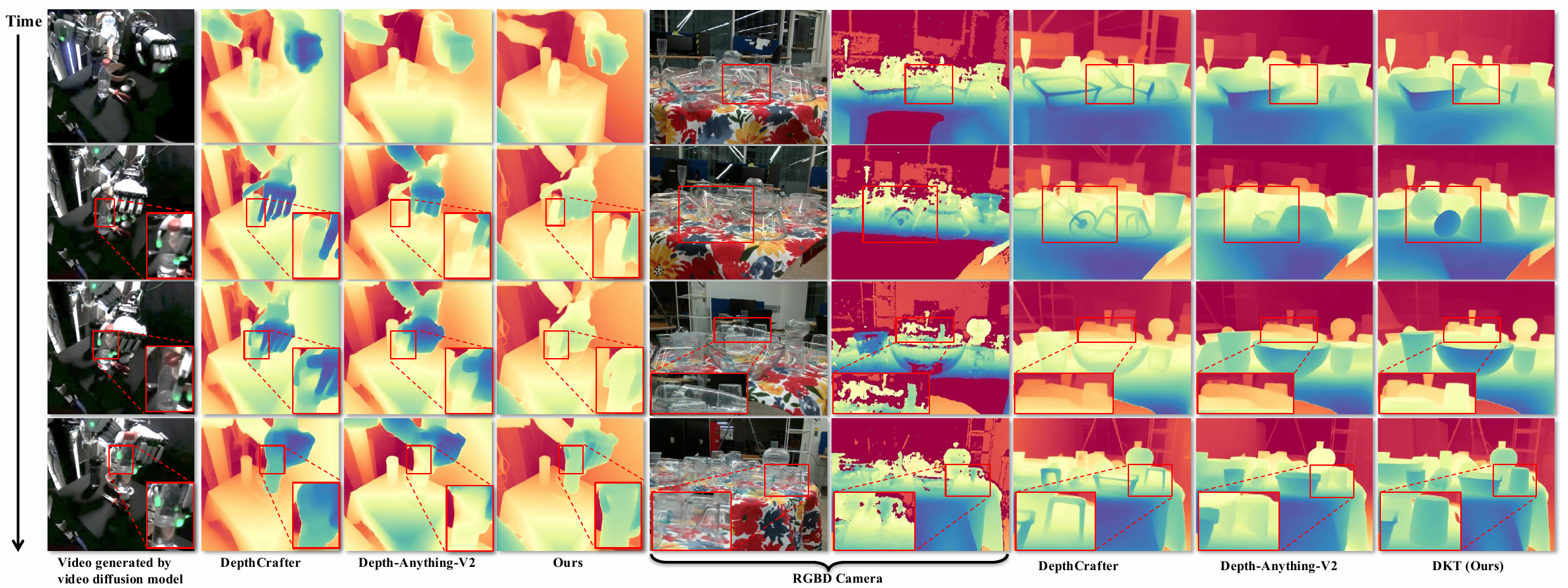}
\caption{We present DKT, a foundational model for fine-grained, temporally consistent depth estimation of in-the-wild videos featuring transparent objects of arbitrary lengths.}
\label{fig:teaser}
\end{figure*}

With recent advances in Video Diffusion Model (VDM) \cite{deepmind2024veo, wan2025}, we observe their remarkable capacity to synthesize physically plausible videos of interactions with transparent objects, as illustrated in the first column of Fig.~\ref{fig:teaser}. Our central insight is that these models appear to have implicitly internalized the physical principles of light transport—such as refraction and reflection through transparent or translucent materials. To leverage this knowledge for video depth estimation of transparent-object, we make contributions from two perspectives: data and learning. 

\textbf{Data.} We collect a 3D asset collection consisting of diverse categories and shapes of transparent and highly reflective items. Subsequently, we introduce a rendering pipeline that automatically generates physically plausible scenes using these assets and renders video data with varied light sources and camera trajectories, leading to the first synthetic video dataset, termed \textit{TransPhy3D}, which focuses on transparent-objects, complementing existing image counterparts \cite{wei2024d,dai2022domain,sajjan2020clear} that primarily study single-frame depth estimation problem. \textbf{Learning.} We propose a paradigm shift to video depth estimation: reframing it from a discriminative estimation task to a video-to-video translation problem. We achieve this by repurposing VDM~\cite{wan2025} using a LoRA training strategy. 
To fully leverage existing frame-wise datasets, we introduce a co-training strategy that enables joint training on a mixture of frame-wise and video data.
Finally, we introduce a foundation model designed primarily for video depth estimation of transparent-object, termed \textbf{DKT}.

We validate the effectiveness of DKT through comprehensive experiments, demonstrating that it achieves SOTA performance under zero-shot setting on both synthetic and real-world benchmarks. 

In summary, our main contributions are:

\begin{itemize}

\item We introduce TransPhy3D, the first synthetic transparent-object video dataset, comprising 11,000 videos and 1.32 million frames of data, to enable effective fine-tuning of VDM.

\item We introduce the first foundation model for transparent-object video depth estimation by repurposing VDM through LoRA finetuning and we design a co-training strategy for training on mixture data of available synthetic image datasets and TransPhy3D.

\item We conduct comprehensive benchmarking of existing SOTA methods on several open datasets and demonstrate the superiority of DKT in both depth estimation accuracy and real-world robotic experiments.
\end{itemize}

\section{RELATED WORKS}

\subsection{From Discriminative to Generative Depth Estimation}

Depth estimation—particularly for video and transparent objects—has long been dominated by discriminative approaches~\cite{wofk2019fastdepth,okae2021robust,li2023bridging,wei2024d,hu2025depthcrafter,wofk2019fastdepth,li2024radarcam,liao2017parse,shivakumar2019real,pillai2019superdepth,wang2024toward,ebner2024metrically,wang2024toward,li2024stereo}. Early methods such as FastDepth~\cite{wofk2019fastdepth} and ClearGrasp~\cite{sajjan2020clear} relied on synthetic data and hand-crafted geometric cues (e.g., surface normals, occlusion boundaries) to overcome the ambiguous visual appearance of transparency. Despite their innovation, these models suffered from significant domain gaps and limited generalization. Subsequent work incorporated stereo cues~\cite{li2024stereo,li2023bridging}, probabilistic volumetric representations~\cite{li2024one}, and metric learning techniques~\cite{ebner2024metrically,wang2024toward}, yet still operated within a discriminative framework, attempting to directly map pixels to depth.
A turning point emerged with the adoption of generative models—especially diffusion models—which reframe depth estimation not as regression, but as a conditional generative process. Methods such as D3RoMa~\cite{wei2024d} and VPD~\cite{li2024one} leverage diffusion to iteratively refine depth predictions, incorporating physical constraints such as stereo consistency and temporal smoothness. These approaches implicitly learn optical priors, enabling more robust inference on challenging materials. More recent video-depth techniques—including RollingDepth~\cite{ke2025video}, FlashDepth~\cite{chou2025flashdepth}, and DepthCrafter~\cite{hu2025depthcrafter}—further demonstrate that generative architectures inherently capture scene dynamics and material properties, even under transparency.
Our work builds upon this generative turn, but goes further: we show that large-scale video diffusion models, pre-trained on internet-scale video data, already internalize a rich prior of transparent phenomena. By fine-tuning such a model \textit{entirely on synthetic data}, we achieve SOTA zero-shot depth estimation without any real-world labels.

\subsection{The Rise of Generative Data and Physics-Aware Synthesis}

Parallel to advances in model architecture, the synthesis of training data has also undergone a generative revolution~\cite{sajjan2020clear,wei2024d,dai2022domain,ni2024ti2v,li2025uniscene,chi2023diffusionpolicy,wen2024vidman,agrawal2024clear,cheng2024gam,dai2025depth, yang2025orv,wang2025semantic,wang2025unifying,ji2017surfacenet,liao2022kitti,zheng2022image,liang2025diffusion,liu2023neudf,xie2018tempogan,lin2025prompting,jiang2024gaussianshader,liu2013fast,chang2024gaussreg,ye2025hi3dgen,chen2018ps,chen2019self,xiong2025octfusion,tao2024maniskill3,chen2019learning,he2024lotus,huang2018holistic,xu2025stable}. Early synthetic datasets for transparency, such as those produced by ClearGrasp~\cite{sajjan2020clear} and DREDS~\cite{dai2022domain}, relied on physically-based rendering (PBR) and careful domain randomization. While effective, these methods required significant expertise and computational resources to simulate realistic sensor noise and material variations.
The advent of generative video models has introduced a new paradigm: models such as TI2V-Zero~\cite{ni2024ti2v} can produce photorealistic, temporally coherent videos of transparent objects \textit{without} fine-tuning, implying that generative priors encapsulate complex optical laws. Subsequent tools like LayerDiffuse~\cite{zhang2024transparent} and Clear-Splatting~\cite{agrawal2024clear} further enable fine-grained control over transparency effects in generated content. These capabilities suggest that generative models have learned not only appearance, but underlying physical rules.
In robotics, this shift has enabled policies and perception systems that leverage generative world models, as seen in Diffusion Policy~\cite{chi2023diffusionpolicy} and VidMan~\cite{wen2024vidman}. Unlike traditional simulation-based data generation, generative models offer scalability and diversity, reducing reliance on hand-engineered graphics pipelines.
Our approach fully embraces this idea: we use a generative video diffusion model as both a data synthesizer and a perception backbone. By fine-tuning it on purely synthetic transparent sequences, we exploit its inherent physical understanding—achieving superior generalization without real-world supervision. 
\section{METHOD}

\begin{figure*}[t]
\includegraphics[width=2\columnwidth]{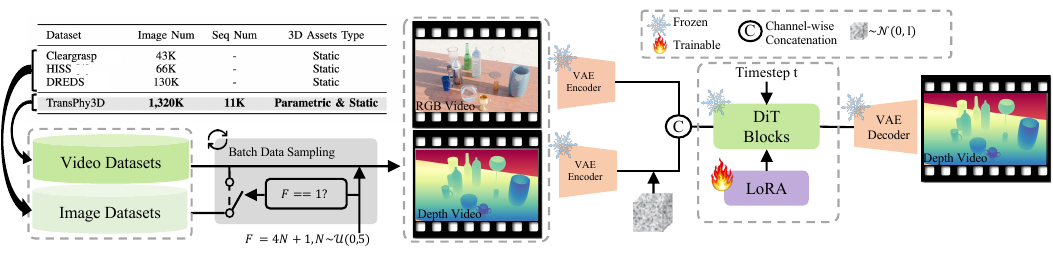}
\caption{\textbf{Overaview of DKT.} DKT starts with a pretrained video diffusion model~\cite{wan2025} and is finetuned for video depth estimation  by concatenating an extra RGB latent with the input latent using LoRA training strategy.}
\label{fig:main}
\end{figure*}


Despite the rapid progress in video depth and normal estimation~\cite{wofk2019fastdepth, ye2024stablenormal, hu2025depthcrafter, bin2025normalcrafter}, existing methods remain limited when handling transparent and reflective objects. The core difficulty lies in the absence of reliable supervision: for real-world data, the SOTA pipeline~\cite{chen2022clearpose}—RGB-D capture followed by CAD model recovery—fails to account for background information, while for synthetic data, no video dataset with transparent objects is available. This scarcity of ground truth highlights the need for strong generative priors. However, adapting generative prior models such as VDMs to this specific domain introduces a critical challenge—catastrophic forgetting of their original priors. 

To tackle these challenges, we propose DKT, a framework that couples large-scale video data curation of transparent and reflective objects with LoRA-based adaptation of VDMs.

To mitigate the lack of supervision, we first construct a 3D asset bank with diverse categories and shapes, and introduce a rendering pipeline that generates physically plausible video scenes under varied lighting and camera trajectories. To further reduce rendering costs and enhance training efficiency, we incorporate existing synthetic image datasets and devise a heuristic sampling strategy that enables joint training on both image and video data within a unified pipeline. To address the scenario of catastrophic forgetting, we employ the LoRA strategy~\cite{hu2022lora} to efficiently adapt VDM for transparency perception, achieving a seamless fusion of its transparent priors with the essential knowledge required for new tasks.

\subsection{TransPhy3D}

To address the gap in video datasets featuring transparent and reflective objects, we construct the first synthetic video dataset of transparent and reflective objects. This dataset is characterized by diversity in object shapes and categories, varied camera trajectories, and high-quality annotations.

\begin{figure}[t]
\includegraphics[width=1\columnwidth]{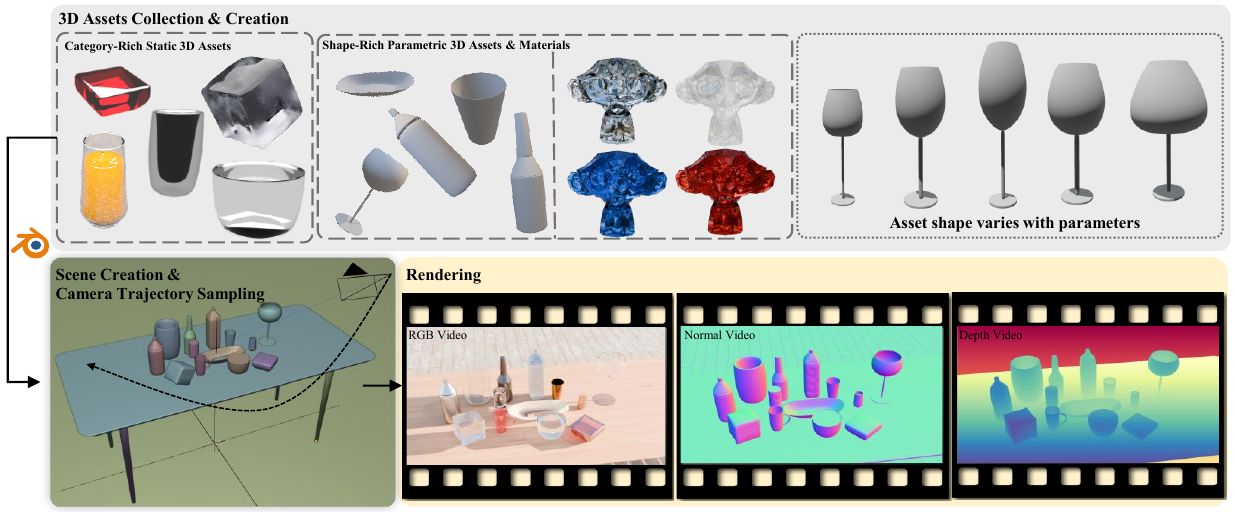}
\caption{\textbf{Rendering Pipeline.} Scenarios are constructed using static and parametric 3D assets. RGB, depth, and normal videos are rendered by sampling a circular trajectory within the scene. }
\label{fig:dataset}
\end{figure}

\textbf{Parametric \& Static 3D Assets.}  As illustrated in Fig.~\ref{fig:dataset}, our asset repository integrates two complementary sources: \textit{Category-Rich Static 3D Assets} and \textit{Shape-Rich Parametric 3D Assets}, ensuring rich categories and shapes. For the former, we collected 5,574 assets from BlenderKit\footnote{\url{https://www.blenderkit.com/}}. Each asset is assigned an aesthetic score by rendering an image and passing it through Qwen2.5-VL-7B~\cite{qwen25} to identify objects with transparent or highly reflective properties. Consequently, this process results in a final collection with 574 high-quality assets that is rich in categories, featuring transparent and reflective assets. For the latter, following \cite{kim2024t2sqnet}, we develop a procedural pipeline to generate parametric assets. As shown in the bottom of Fig.\ref{fig:dataset}, varying parameters of the same asset can produce different shapes. Consequently, this procedure yields a collection that is rich in shape diversity. 

To give these models a photorealistic appearance, we pair them with a specially curated material library containing a wide selection of transparent materials (like glass and plastic) and highly reflective ones (like metal and glazed ceramic).

\textbf{Scene Creation.} This stage focuses on composing scenes dynamically through physics simulation. we randomly select $M$ assets and initialize their six-degree-of-freedom (6-DOF) poses and scales within a predefined environment, such as a container or tabletop, as shown in the topright of Fig.~\ref{fig:dataset}, We then employ Blender's integrated physics engine to simulate the objects as they fall and collide, allowing them to settle into a physically plausible and natural final arrangement.

\textbf{Camera Sampling \& Rendering.} To capture diverse and dynamic viewpoints, our camera sampling method generates circular trajectories around the geometric center of objects, incorporating sinusoidal perturbations of varying amplitudes. We then utilize Blender's ray tracing engine, Cycles, to perform physically accurate lighting calculations and material rendering. This process precisely simulates complex light transport phenomena, including propagation, refraction, and reflection within transparent materials. As a final step, we use the NVIDIA OptiX-Denoiser to optimize image quality.

The output of this pipeline is \textbf{TransPhy3D}, a novel video dataset comprising \textbf{11,000} unique scenes. With each scene rendered as a 120-frame video, the dataset contains a total of \textbf{1,320,000} frames, sourced from both our parametric and static asset collections.




\subsection{Preliminaries of Video Diffusion Model}
This work builds upon WAN~\cite{wan2025}, which is comprised of three primary components: a VAE, a diffusion transformer consisting of multiple DiT blocks, and a text encoder. The VAE compresses input videos into latent space and decodes predicted latents back to image space. The text encoder encodes text prompts into embeddings. The diffusion transformer predicts velocity given noisy latents and text embeddings.

WAN leverages the flow matching framework~\cite{lipman2024flow} to model a unified denoising diffusion process. During training, given an image or video latent $\mathbf{x}_1$, a random noise $\mathbf{x}_0 \sim \mathcal{N}(\mathbf{0}, \mathbf{I})$, and a timestep $\mathbf{t} \sim \mathcal{U}(0,1)$, an intermediate latent $\mathbf{x}_t$ serving as training input is obtained by:
\begin{equation}
    \mathbf{x}_t = \mathbf{t} \mathbf{x}_1 + (1 - \mathbf{t})\mathbf{x}_0.
    \label{eq:diffusion}
\end{equation}
The ground truth velocity $\mathbf{v}_t$ is obtained by:
\begin{equation}
    \mathbf{v}_\mathbf{t} =  \frac{\mathbf{dx}_\mathbf{t}}{\mathbf{dt}}= \mathbf{x}_1 - \mathbf{x}_0.
    \label{eq:gt_construction}
\end{equation}
The loss function is MSE between the output of a velocity predictor $\mathbf{u}$ and $\mathbf{v}_t$:
\begin{equation}
    \mathcal{L} = \mathbb{E}_{\mathbf{x}_0, \mathbf{x}_1, \mathbf{c}_\text{txt}, \mathbf{t}}
        \bigg{ \Vert} \mathbf{u} \big{(} \mathbf{x}_\mathbf{t}, \mathbf{c}_\text{txt}, \mathbf{t} \big{)} - \mathbf{v}_t \bigg{ \Vert}^2,
    \label{eq:flow_matching_loss}
\end{equation}
where $\mathbf{c}_\text{txt}$ is the text embedding.

\begin{figure*}[ht]
\includegraphics[width=1.95\columnwidth]{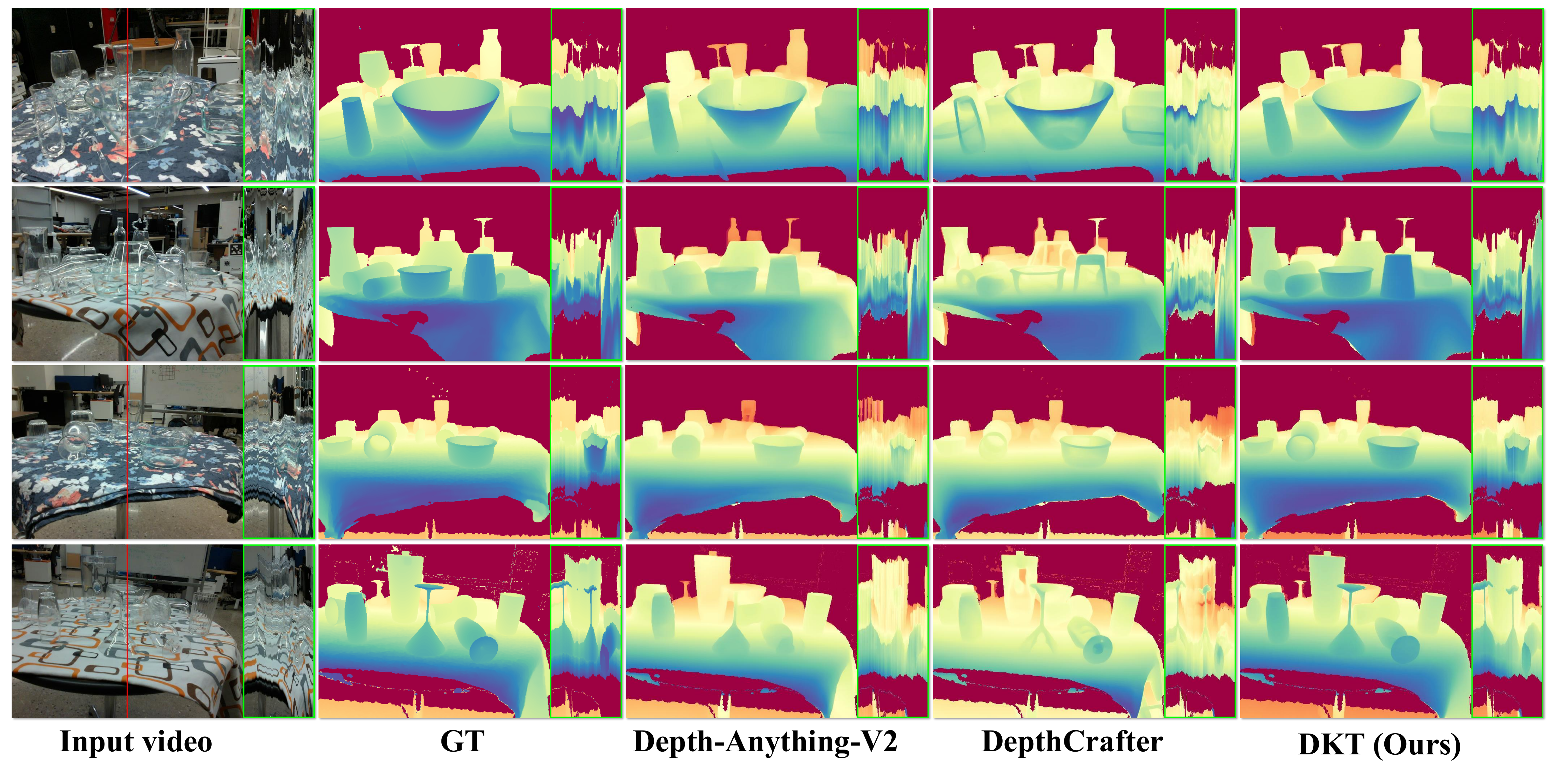}
\caption{\textbf{Qualitative comparison on the ClearPose~\cite{chen2022clearpose}.} For better visualizing the temporal quality, we show the temporal profiles of each result in green boxes, by slicing the depth values along the time axis at the red line positions. }
\label{fig:qualitative_clearpose_depth}
\end{figure*}

\subsection{Training Strategy }

As illustrated in Fig.~\ref{fig:main}, to enhance the training efficiency and alleviate the rendering burden of rendering process.  We propose to co-train synthetic image and video data (TransPhy3D). 

We first sample a constant number $F$ using: 
\begin{equation}
     \begin{aligned}
          F &= 4N + 1 \\  
          N &\ \sim \mathcal{U}(0,5).
     \end{aligned}
\end{equation}
$F$ indicates the frame number for the video in this batch of data. Afterwards, if $F$ is equal to $1$, we sample a batch of paired data consisting of RGB and depth videos from both video and image datasets (with the video containing only one frame); otherwise, we sample only from video datasets.


Afterwards, the overall pipeline of DKT is presented in Fig.~\ref{fig:main}, the depth video in one pair is converted to disparity. Both RGB and depth videos are normalized to $[-1,1]$ to match the VAE training space, then encoded by the VAE into latents $\mathbf{x}^c_1$ and $\mathbf{x}^d_1$.

The depth latent $\mathbf{x}^d_1$ is transformed by Eq.~\ref{eq:diffusion} into the intermediate latent $\mathbf{x}^d_\mathbf{t}$. The input to the DiT blocks is then obtained by concatenating $\mathbf{x}^d_\mathbf{t}$ and $\mathbf{x}^c_1$ along the channel dimension. The training loss is defined as the difference between the DiT output and the ground-truth velocity $\mathbf{v}^d_t$ (constructed by Eq.~\ref{eq:gt_construction}), and is computed as:
\begin{equation}
    \mathcal{L} = \mathbb{E}_{\mathbf{x}_0, \mathbf{x}^d_1,\mathbf{x}^c_1, \mathbf{c}_\text{txt}, \mathbf{t}}
        \bigg{ \Vert} \mathbf{u} \big{(} \textbf{Concat}(\mathbf{x}^d_\mathbf{t},\mathbf{x}^c_\mathbf{1}), \mathbf{c}_\text{txt}, \mathbf{t} \big{)} - \mathbf{v}^d_t \bigg{ \Vert}^2,
    \label{eq:flow_matching_loss}
\end{equation}
where $\mathbf{c}_\text{txt}$ is the text embedding, and \textbf{Concat} denotes the concatenation operation along the channel dimension.

All model components remain frozen except for a small set of trainable LoRA \cite{hu2022lora} parameters in the DiT, which learn low-rank weight adaptations.

\begin{table*}[!ht]
    \centering
    \caption{Quantitative comparison for video depth estimation on ClearPose and TransPhy3D-Test. \textbf{Best} and \underline{second best} are highlighed.}
    \vspace{-10px}
    \resizebox{1.9\columnwidth}{!}{
    \begin{tabular}{lcccccccccccc}
    \toprule
        \multirow{2}{*}{Methods} & \multicolumn{5}{c}{\textbf{ClearPose \cite{chen2022clearpose}}} & \multicolumn{5}{c}{\textbf{TransPhy3D-Test}} \\ 
        
         \cmidrule(lr){2-7} \cmidrule(lr){8-13} & $\text{REL}\downarrow$ & $\text{RMSE}\downarrow$ & $\delta_{1.05}\ \uparrow$  & $\delta_{1.10}\ \uparrow$  & $\delta_{1.25}\ \uparrow$  & Rank $\downarrow$& $\text{REL}\downarrow$ & $\text{RMSE}\downarrow$ & $\delta_{1.05}\ \uparrow$ & $\delta_{1.10}\ \uparrow$ & $\delta_{1.25}\ \uparrow$ & Rank $\downarrow$ \\ 
        \midrule
        
        Depth4ToM~\cite{costanzino2023learning} (ICCV23) & 12.38  & 14.02  & 28.74 & 51.39 & 85.94 & 4.0 & 18.01  & 720.13  & 31.54 & \underline{56.56} & 85.98 & 3.8 \\ 
        
        DAv2~\cite{yang2024depth} (NeurIPS24) & \underline{10.85}  & \textbf{12.21}  & \underline{32.21} & \underline{56.37} & \underline{89.94} & \textbf{1.8}
        & 14.02  & 74.86  & 31.36 & 53.12 & 81.27 & 4.0 \\ 
        
        Marigold-E2E-FT~\cite{martingarcia2024diffusione2eft} (WACV25)& 16.44  & 16.65  & 16.99 & 33.85 & 74.24 & 7.0 & 23.58  & 42.63  & 11.42 & 27.49 & 62.86 & 5.4 \\ 
        
        MoGe~\cite{wang2025moge} (CVPR25) & 13.13  & 13.40  & 24.09 & 45.08 & 84.29 & 4.6 & 31.91  & 136.74  & 19.29 & 32.24 & 50.98 & 5.8 \\

        VGGT~\cite{wang2025vggt} (CVPR25) & 15.38  & 15.68  & 19.93 & 38.33 & 76.89 & 6.0 & 32.30  & 49.82  & 13.64 & 30.13 & 55.25 & 5.8 \\ 

        DepthCrafter~\cite{hu2025depthcrafter} (CVPR25) & 11.32  & \underline{12.34}  & 31.92 & 55.46 & 88.59 & \underline{2.8} & \underline{11.32}  & \textbf{12.34}  & \underline{31.92} & 55.46 & \underline{88.59} & \underline{2.0} \\ 
        
        \midrule
        DKT (Ours) & \textbf{9.72} & 14.58  & \textbf{38.17} & \textbf{65.50} & \textbf{93.04} & \textbf{1.8} & \textbf{2.96}  & \underline{19.50}  & \textbf{87.17} & \textbf{97.09} & \textbf{98.56} & \textbf{1.2} \\ 
        \bottomrule
        
    \end{tabular}}
    \label{tab:clearpose_transPhy3d}
    \vspace{-8px}
\end{table*}

\begin{table*}[!ht]
    \centering
    \caption{Quantitative comparison for video depth estimation on the DREDS datasets.}
    \resizebox{1.9\columnwidth}{!}{
    \begin{tabular}{lcccccccccccc}
    \toprule
        \multirow{2}{*}{Methods} & \multicolumn{5}{c}{\textbf{DREDS-STD-CatKnown \cite{dai2022domain}}}  & \multicolumn{5}{c}{\textbf{DREDS-STD-CatNovel \cite{dai2022domain}}}  \\ 
         \cmidrule(lr){2-7} \cmidrule(lr){8-12} & $\text{REL}\downarrow$ & $\text{RMSE}\downarrow$ & $\delta_{1.05}\ \uparrow$ & $\delta_{1.10}\ \uparrow$ & $\delta_{1.25}\ \uparrow$ & Rank$\downarrow$ & $\text{REL}\downarrow$ & $\text{RMSE}\downarrow$ & $\delta_{1.05}\ \uparrow$ & $\delta_{1.10}\ \uparrow$ & $\delta_{1.25}\ \uparrow$ & Rank$\downarrow$ \\ 
        \midrule
        
        Depth4ToM~\cite{costanzino2023learning} (ICCV23)& 6.92  & 6.60  & 44.23 & 74.15 & 98.19 & 5.6 & 7.21  & 5.77  & 43.45 & 71.13 & 98.04 & 5.6 \\ 
        
        DAv2~\cite{yang2024depth} (NeurIPS24) &  \underline{6.94}  & 6.58  & 44.46 & 73.66 & 98.32 & 5.4 & 7.41  & 5.76  & 41.16 & 69.77 & 98.05 & 6.2 \\ 

        Marigold-E2E-FT~\cite{martingarcia2024diffusione2eft} (WACV25) & 6.07  & 5.81  & 49.07 & 79.15 & 99.36 & 3.2 & 7.06  & 5.55  & 42.81 & 71.51 & 98.71 & 4.4 \\
        
        MoGe~\cite{wang2025moge} (CVPR25) & 6.95  & \underline{5.78}  & \underline{47.03} & \underline{74.44} & 97.46 & 4.8 & \underline{6.07}  & \textbf{4.32}  & \underline{50.25} & \underline{78.95} & \underline{99.31} & \underline{2.0} \\ 
        
        VGGT~\cite{wang2025vggt} (CVPR25) & 5.74  & 5.14  & 51.14 & 80.94 & 99.79 & \underline{2.0} & 6.10  & 4.74  & 48.93 & 78.56 & 99.53 & 2.8 \\ 

        DepthCrafter~\cite{hu2025depthcrafter} (CVPR25) & 7.06  & 6.41  & 41.45 & 72.32 & \underline{98.68} & 6.2 & 7.41  & 5.54  & 38.44 & 70.23 & 98.51 & 5.6 \\

        \midrule
        DKT (ours) & \textbf{5.30}  & \textbf{4.96}  & \textbf{53.86} & \textbf{84.93} & \textbf{99.89} & \textbf{1.0}  & \textbf{5.71}  & \underline{4.66}  & \textbf{52.12} & \textbf{79.51} & \textbf{99.84} & \textbf{1.2}  \\ 
        \bottomrule
    \end{tabular}}
    \label{tab:dreds}
    \vspace{-10px}
\end{table*}


\subsection{Implementation}

We trained our model with a learning rate of $1e-5$ using AdamW~\cite{loshchilov2017decoupled} and a batch size of 8. The model is trained using synthetic image datasets, including HISS~\cite{wei2024d}, DREDS~\cite{dai2022domain}, and ClearGrasp~\cite{sajjan2020clear}, along with our newly introduced video synthetic dataset, TransPhy3D. Following the training strategy of WAN, all datasets are resized to $832 \times 480$ for model training. The number of training iterations is 70K, which takes 8 Nvidia H100 GPUs for two days. Except for specific explanations, the denoising step is set to 5 for inference. By following the inference strategy in ~\cite{hu2025depthcrafter}, we achieve arbitrary-length video inference by splitting the input into overlapping segments. Consecutive segments are then stitched together using a complementary weight applied to the overlapping regions. For more details, please refer to ~\cite{hu2025depthcrafter}.

\section{EXPERIMENT}



\begin{figure*}[ht]
\includegraphics[width=1.95\columnwidth]{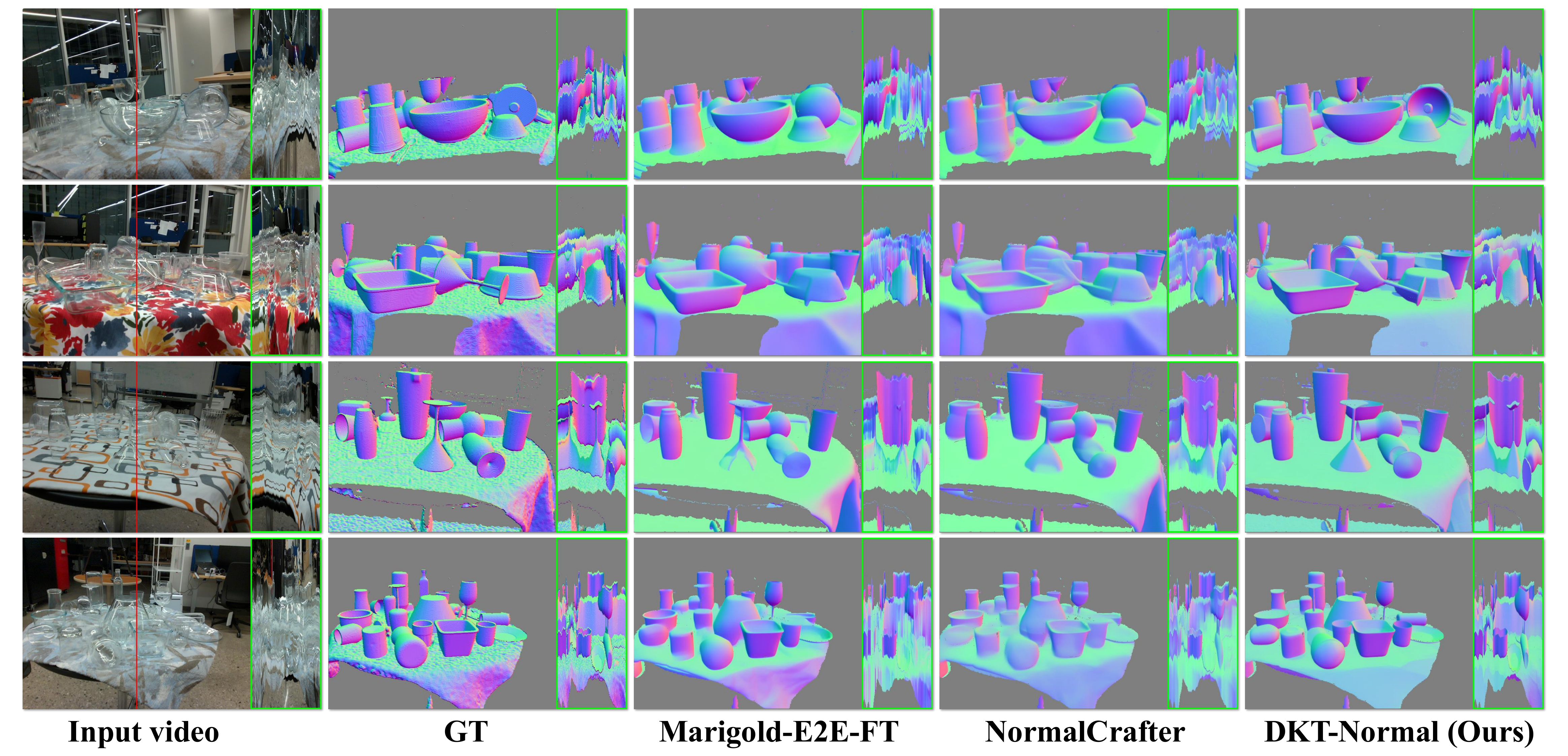}

\caption{\textbf{Qualitative comparison for video normal estimation on ClearPose~\cite{chen2022clearpose}.}}
\label{fig:qualitative_clearpose_normal}

\end{figure*}

\subsection{Evaluation Metrics}


Following the practice of evaluating the temporal consistency of depth maps\cite{hu2025depthcrafter}, the predictions are aligned with the ground truth using a global scale and shift. The following metrics are then calculated: $\delta_{1.05}$, $\delta_{1.10}$, $\delta_{1.25}$, and $\text{REL}$, which are expressed as percentages, along with $\text{RMSE}$, measured in centimeters. 


\subsection{Evaluation Datasets}

DKT is evaluated using the following real-world and synthetic datasets under a \textbf{zero-shot} setting to demonstrate its generalization, robustness, and potential effects on the robotic community.

\textbf{ClearPose}: ClearPose~\cite{chen2022clearpose} is a real-world RGB-D benchmark for transparent and translucent objects. This testset consists of 27 real-world scenes, including different backgrounds, heavy occlusions, objects in translucent and opaque covers, non-planar surfaces, and even scenes filled with liquid. Each scene is captured in a long video using the RealSense L515 depth camera. The foreground ground truth depth of transparent objects is recovered using object CAD models. The background region is masked by a threshold of $[0.3, 1.5]$ during evaluation. This dataset also provides normal annotation translated from depth map.

\textbf{STD}: STD~\cite{dai2022domain} is a real-world dataset comprising specular, transparent, and diffuse objects. It is divided into two subsets, CatKnown and CatNovel, based on the commonality of the objects. The former comprises 12 scenes, while the latter contains 5 scenes.

\textbf{TransPhy3D-Test}: We employ a new transparent model~\cite{kim2024transpose} to render a synthetic test set using the pipeline we introduce. This dataset consists of 28 scenes.

\subsection{State-of-the-art Comparision}

A comparison is conducted with available foundation depth estimation methods, including the image depth estimation method Depth-Anything-V2~\cite{yang2024depth}, MoGe~\cite{wang2025moge}, VGGT~\cite{wang2025vggt}, Marigold-E2E-FT~\cite{martingarcia2024diffusione2eft}, Depth4ToM~\cite{costanzino2023learning} and the video depth estimation method DepthCrafter~\cite{hu2025depthcrafter}.

As shown in Tab.~\ref{tab:clearpose_transPhy3d} and~\ref{tab:dreds}, DKT sets a new SOTA across three real-world datasets and one synthetic dataset. The performance gap peaks in ClearPose and TransPhy3D, both of which exclusively involve transparent and highly reflective objects. Specifically, we outperform the second-best method by scores of $5.69$, $9.13$, and $3.1$ for $\delta_{1.05}$, $\delta_{1.10}$, and $\delta_{1.25}$ in ClearPose, and $55.25$, $40.53$, and $9.97$ for $\delta_{1.05}$, $\delta_{1.10}$, and $\delta_{1.25}$ in TransPhy3D.

Moreover, this advantage is clearly reflected in Fig.~\ref{fig:qualitative_clearpose_depth}. Following ~\cite{hu2025depthcrafter}, we present a temporal profile to better illustrate the temporal consistency of the predicted video depth. DKT not only achieves superior identification of transparent objects in the first frame but also demonstrates optimal temporal consistency.

\textbf{What accounts for the significant performance gap in the TransPhy3D-Test?} We attribute this phenomenon to a characteristic of our rendering data: the camera trajectories during rendering describe a circular path around the object. This feature increases the requirements for inter-frame consistency in the model's predictions, as even a minor error can lead to a significant decline in prediction accuracy after global alignment.

\subsection{Ablation Study}

\begin{table}[t]
    \centering
    \caption{Ablation for training strategy in ClearPose. $\times$ indicates naive finetuning.}
    \resizebox{0.95\columnwidth}{!}{
    \begin{tabular}{llccccc}
    \toprule
        Model Size & LoRA & $\text{REL}\downarrow$ & $\text{RMSE}\downarrow$ & $\delta_{1.05}\ \uparrow$  & $\delta_{1.10}\ \uparrow$  & $\delta_{1.25}\ \uparrow$  \\
         \midrule
        1.3B & $\times$ & 11.86  & 26.54  & 30.48 & 54.03 & 88.30 \\ 
        1.3B & $\checkmark$ & 11.17  & 17.45  & 33.16 & 58.02 & 90.65 \\ 
        14B & $\checkmark$ & \textbf{9.72}  & \textbf{14.58}  & \textbf{38.17} & \textbf{65.50} & \textbf{93.04} \\ 
        \bottomrule
    \end{tabular}
    }
    \label{tab:training_strategy}
    \vspace{-10px}
\end{table}
\textbf{Training Strategies.}  As illustrated in Tab.~\ref{tab:training_strategy}, naive finetuning results in high computational costs and suboptimal performance compared to LoRA fine-tuning. By adopting the LoRA training strategy and scaling up the model size, a significant improvement in perforamnce is achieved.

\begin{figure}[t]
\includegraphics[width=1\columnwidth]{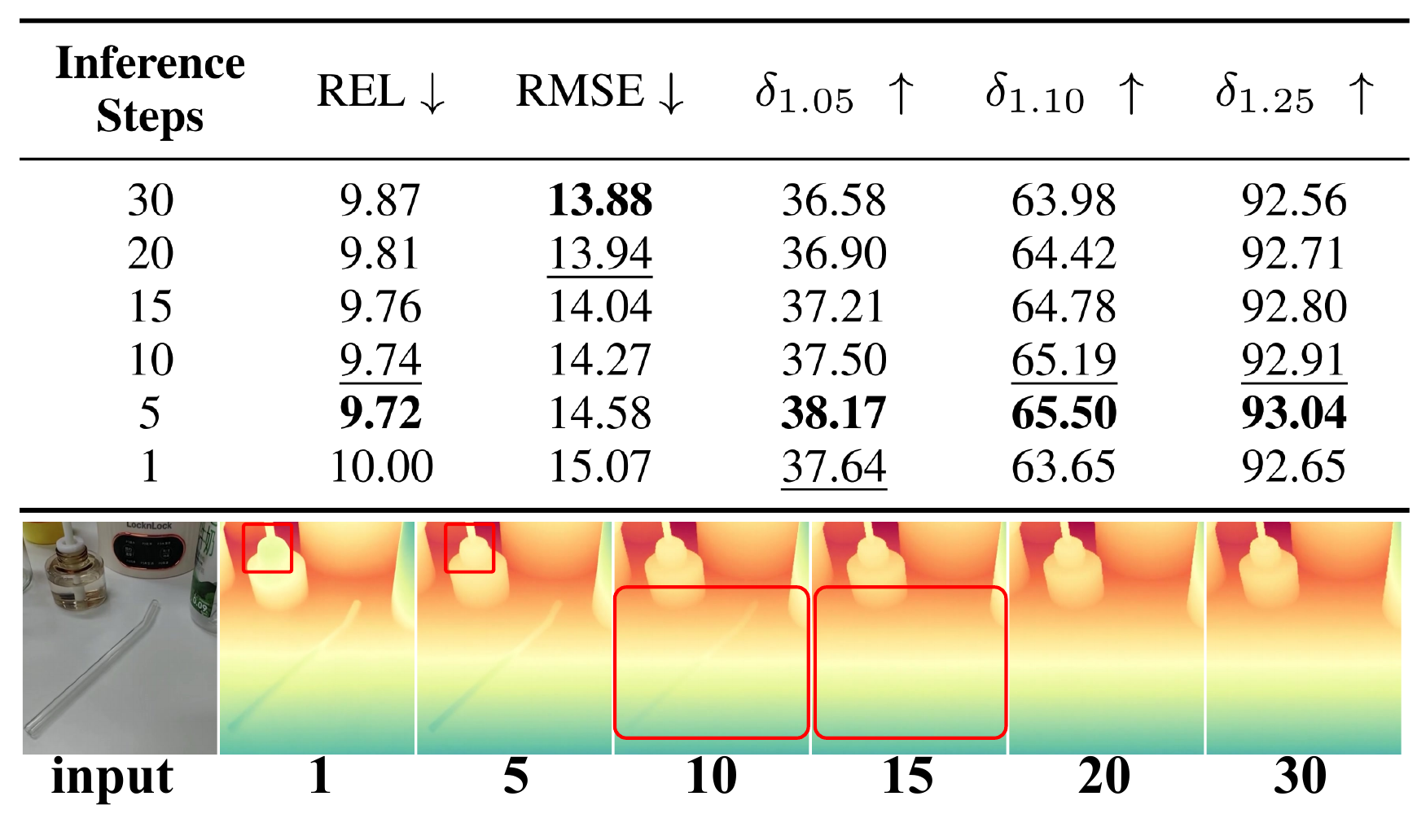}
\caption{\textbf{The effects of different Inference step.}}
\label{fig:inference_step_concept}
\end{figure}

\textbf{Inference Steps} As illustrated in the upper part of Fig.~\ref{fig:inference_step_concept}, increasing the number of inference steps does not yield significant performance improvements. Moreover, as shown in the lower part of Fig.~\ref{fig:inference_step_concept}, fewer steps result in inaccurate predictions, while more inference steps lead to the loss of important details. To balance performance and inference efficiency, we set 5 as the default number of inference steps.

\begin{table}[t]
\caption{Inference time per frame (ms) at a resolution of $832 \times 480$. }
    \resizebox{0.95\columnwidth}{!}{
    \centering
    \begin{tabular}{lcccc}
    \toprule
        \textbf{Method} & \textbf{Encoding} & \textbf{Denoising} & \textbf{Decoding} & \textbf{All} \\ 
        \midrule
        DAv2~\cite{yang2024depth} & N/A & N/A & N/A & 277.75 \\ 
        DepthCrafter~\cite{hu2025depthcrafter} & 141.85 & 240.01 & 183.69 & 565.55 \\ 
        DKT-14B & 46.53 & 297.11 & 68.07 & 411.71 \\ 
        DKT-1.3B & 46.53 & 52.88 & 68.07 & 167.48 \\ 
        \bottomrule
    \end{tabular}
    }
    \label{tab:inference_efficiency}
    \vspace{-10px}
\end{table}
\textbf{Computational Efficiency} To assess inference efficiency fairly, we reevaluated DKT, the baseline DAv2-Large~\cite{yang2024depth}, and DepthCrafter~\cite{hu2025depthcrafter} on the Nvidia L20. The results, shown in Tab.~\ref{tab:inference_efficiency}, indicate that DKT-1.3B achieves the highest efficiency, with a remarkable inference time of only 167.48ms per frame, surpassing DAv2 by 110.27ms. Moreover, the peak GPU memory occupancy of DKT-1.3B is only 11.19 GB, which is acceptable for most robot computational platforms.

\begin{table}[t]
    \centering
    \caption{Video normal estimation on the ClearPose.}
    \resizebox{0.95\columnwidth}{!}{
    \begin{tabular}{lccccc}
    \toprule
        ~ & mean$\downarrow$ & med$\downarrow$ & 11.25$^\circ\uparrow$ & 22.5$^\circ\uparrow$ & 30$^\circ\uparrow$ \\ 
        \midrule
        NormalCrafter~\cite{bin2025normalcrafter} (ICCV25) & \underline{27.08}  & 20.29  & 26.10  & 55.37  & 68.81  \\ 
        Marigold-E2E-FT~\cite{martingarcia2024diffusione2eft} (WACV25) & \underline{27.08}  & \underline{19.40}  & \underline{29.78}  & \underline{57.22}  & \underline{69.30}  \\ 
        DKT-Normal-14B & \textbf{26.03}  & \textbf{18.59}  & \textbf{30.06}  & \textbf{59.63}  & \textbf{70.98}  \\ 
    \bottomrule
    \end{tabular}
    }
    \label{tab:normal_compare}
    \vspace{-10px}
\end{table}

\begin{figure}[htb]
\includegraphics[width=0.90\columnwidth]{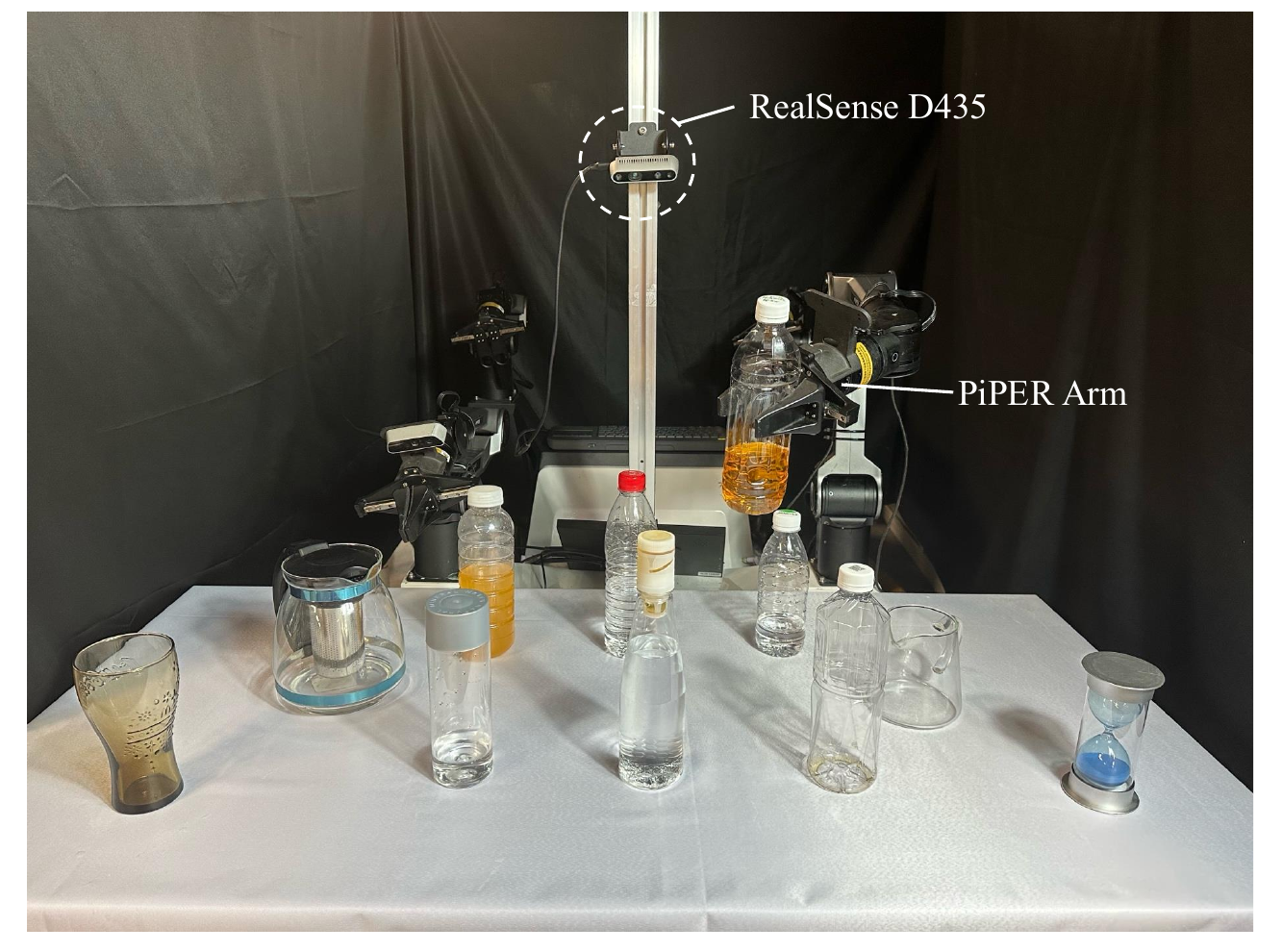}
\caption{\textbf{Real-World Setup.} We use the Cobot Magic system, which integrates the PiPER Arm and RealSense D435 for our grasping tasks.}
\label{fig:robot_setup}
\end{figure}

\begin{figure}[t]
\includegraphics[width=0.95\columnwidth]{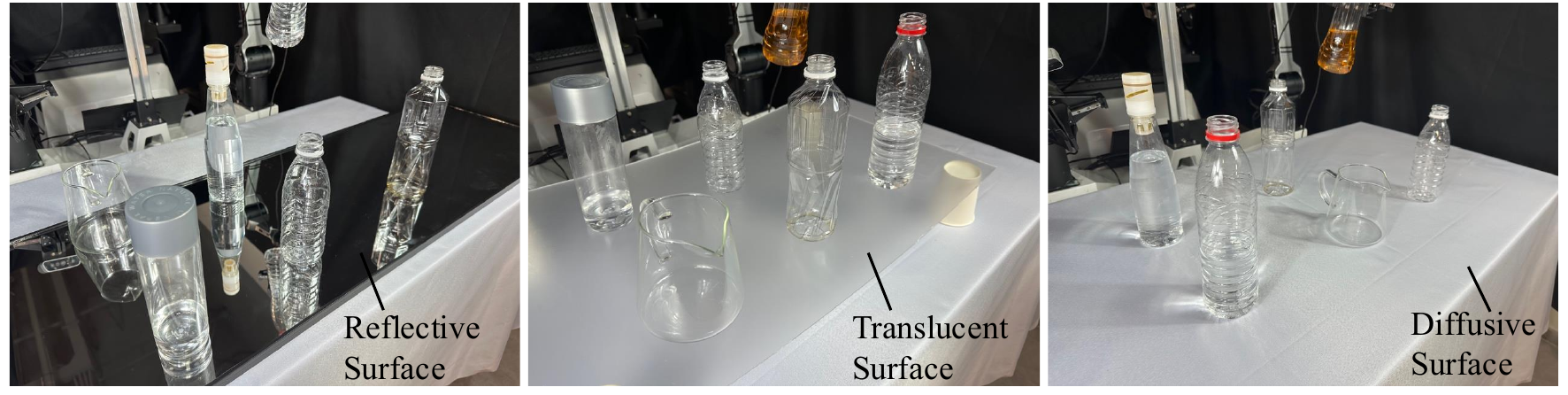}
\caption{\textbf{Demonstration of Surface Types.}}
\label{fig:surface_concept}
\end{figure}


\textbf{Video Normal Estimation} To further validate the efficacy of our training strategy, DKT-Normal-14B is introduced following the same training strategy of DKT-14B. As demonstrated in Tab.~\ref{tab:normal_compare}, DKT-Normal-14B significantly outperforms the previous SOTA video normal estimation method NormalCrafter \cite{bin2025normalcrafter} and the previous SOTA image estimation method Marigold-E2E-FT \cite{martingarcia2024diffusione2eft} by a substantial margin on the zero-shot dataset ClearPose, using the same metrics as NormalCrafter. Moreover, DKT-Normal-14B achieves the sharpest normal prediction and the highest temporal consistency, as illustrated in Fig.~\ref{fig:qualitative_clearpose_normal}.


\begin{table}[t]
    \centering
    \caption{\textbf{Grasping success rate of different depth estimators on translucent, reflective, diffusive surface, respectively.} }
    \begin{tabular}{lcccc}
    \toprule
        Method & Translucent & Reflective & Diffusive & Mean \\ 
        \midrule
        RAW & 0.47 & 0.18 & 0.56 & 0.384 \\ 
        DAv2 & 0.60 & \underline{0.27} & 0.56 & 0.46 \\ 
        DepthCrafter & \underline{0.67} & 0.23 & \underline{0.625} & \underline{0.48} \\ 
        DKT-1.3B & \textbf{0.80} & \textbf{0.59} & \textbf{0.81} & \textbf{0.73} \\ 
        \bottomrule
    \end{tabular}
    \label{tab:robot_manipulation_exp}
    \vspace{-10px}
\end{table}

\subsection{Real-world Grasping Experiments}



\textbf{Experimental Setup. } \textit{1) Hardware System:} As shown in Fig.~\ref{fig:robot_setup}, our real-world robotic manipulation environment consists of two PiPER ARM manipulators used to grasp objects and a fixed-view Realsense D435 camera to provide RGB observations.
\textit{2) Environment Arrangement:} As illustrated in Fig.~\ref{fig:surface_concept}, we set up three types of tabletop backgrounds: reflective, translucent, and diffusive surfaces. Various objects with specular, transparent, and diffusive properties are placed on the table. This setup allows for a comprehensive evaluation of the effectiveness and robustness of different methods under complex real-world scenarios.

\textbf{Deployment Pipeline. } 
Initially, an RGB image is captured by the D435 camera and processed with various relative depth estimation models, including DAv2-Large, DepthCrafter, and DKT-1.3B, to generate relative depth maps. These are then rescaled to metric depth using AprilTag~\cite{wang2016apriltag}. Subsequently, the RGB image and metric depth are input into AnyGrasp~\cite{fang2023anygrasp} to generate a 7-DoF grasp pose (end-effector pose + gripper width). Finally, CuRobo~\cite{sundaralingam2023curobo} is utilized to plan an executable trajectory, which is executed by the PiPER ARM.

\textbf{Results and Analysis.}
As demonstrated in Tab.~\ref{tab:robot_manipulation_exp}, DKT consistently outperforms all baseline across all settings by a significant margin. For the perception results and the grasping video, please refer to the appendix.

\section{CONCLUSIONS}
In this work, we dive into the challenging problem of video depth and normal estimation for transparent and highly reflective scenarios, which is crucial for robotic perception. Our contributions can be summarized in three parts. First, the first video dataset for transparent and highly reflective objects is introduced, accompanied by diverse transparent asset categories and an infinite variety of 3D shapes. Second, DKT is introduced, finetuned from a video diffusion model using a LoRA training strategy. Finally, we demonstrate DKT's performance in comprehensive benchmarks, including synthetic and real-world datasets, and DKT sets a new SOTA across all of them in video depth and normal estimation.

\section{Acknowledgements}
This study is supported by the Beijing Academy of Artificial Intelligence (BAAI), under its research funding programs.
{
\small
\bibliographystyle{ieeetr}
\bibliography{main}
}

\end{document}